\documentclass[]{spie}  

\usepackage{amsmath,amsfonts,amssymb}
\usepackage{graphicx}
\usepackage[colorlinks=true, allcolors=blue]{hyperref}
\usepackage[nolist]{acronym}

\title{Joint tone mapping and denoising of thermal infrared images via multi-scale Retinex and multi-task learning}

\author[a]{Axel G{\"o}drich}
\author[b]{Daniel K{\"o}nig}
\author[c]{Gabriel Eilertsen}
\author[b]{Michael Teutsch}
\affil[a]{Nuremberg Institute of Technology, Nuremberg, Germany}
\affil[b]{Hensoldt Optronics GmbH, Oberkochen, Germany}
\affil[c]{Link{\"o}ping University, Sweden}

\authorinfo{Author e-mail: axel.goedrich@th-nuernberg.de, \{daniel.koenig, michael.teutsch\}@hensoldt.net, gabriel.eilertsen@liu.se}

\begin{document} 
\maketitle

\begin{abstract}
Cameras digitize real-world scenes as pixel intensity values with a limited value range given by the available bits per pixel (bpp). High Dynamic Range (HDR) cameras capture those luminance values in higher resolution through an increase in the number of bpp. Most displays, however, are limited to 8 bpp. Na{\"i}ve HDR compression methods lead to a loss of the rich information contained in those HDR images. In this paper, tone mapping algorithms for thermal infrared images with 16 bpp are investigated that can preserve this information. An optimized multi-scale Retinex algorithm sets the baseline. This algorithm is then approximated with a deep learning approach based on the popular U-Net architecture. The remaining noise in the images after tone mapping is reduced implicitly by utilizing a self-supervised deep learning approach that can be jointly trained with the tone mapping approach in a multi-task learning scheme. Further discussions are provided on denoising and deflickering for thermal infrared video enhancement in the context of tone mapping. Extensive experiments on the public FLIR ADAS Dataset prove the effectiveness of our proposed method in comparison with the state-of-the-art.
\end{abstract}


\begin{acronym}
\acro{HVS}[HVS]{Human Visual System}
\acro{HDR}[HDR]{High Dynamic Range}
\acro{bpp}[bpp]{bits per pixel}
\acro{LDR}[LDR]{Low Dynamic Range}
\acro{TMO}[TMO]{Tone Mapping Operator}
\acro{VIS}[VIS]{Visual-Optical}
\acro{TIR}[TIR]{Thermal Infrared}
\acro{SSIM}{Structural Similarity}
\acro{TMQI}{Tone Mapping Quality Index}
\acro{NSS}{Natural Scene Statistics}
\acro{MWIR}{Medium Wavelength Infrared}
\acro{LWIR}{Long Wavelength Infrared}
\acro{fps}{frames per second}
\acro{CNN}{Convolutional Neural Network}
\acro{DCNN}{Deep Convolutional Neural Network}
\acro{MSE}{Mean Square Error}
\acro{MAE}{Mean Absolute Error}
\acro{MS-SSIM}{Multi-Scale Structural Similarity}
\acro{GAN}{Generative Adversarial Network}
\acro{CAN}{Context Aggregation Network}
\acro{GPU}{Graphics Processing Unit}
\acro{VCN}{Volumetric Correspondence Network}
\acro{MS COCO}{Microsoft Common Objects in Context}
\acro{CLAHE}{Contrast Limited Adaptive Histogram Equalization}
\acro{AHE}{Adaptive Histogram Equalization}
\acro{HE}{Histogram Equalization}
\acro{ODM}{Object Detection Measure}
\acro{mAP}{mean Average Precision}
\acro{CGF}{Conditional Gaussian Filter}
\acro{GT}[GT]{Ground Truth}
\acro{MSR}[MSR]{Multi-Scale Retinex}
\acro{SCB}[SCB]{Simplest Color Balance}
\acro{ML}{Machine Learning}
\end{acronym}

\section{INTRODUCTION}
\label{sec:intro}

\begin{figure}[tbp]
\centering
\includegraphics[width=1\textwidth]{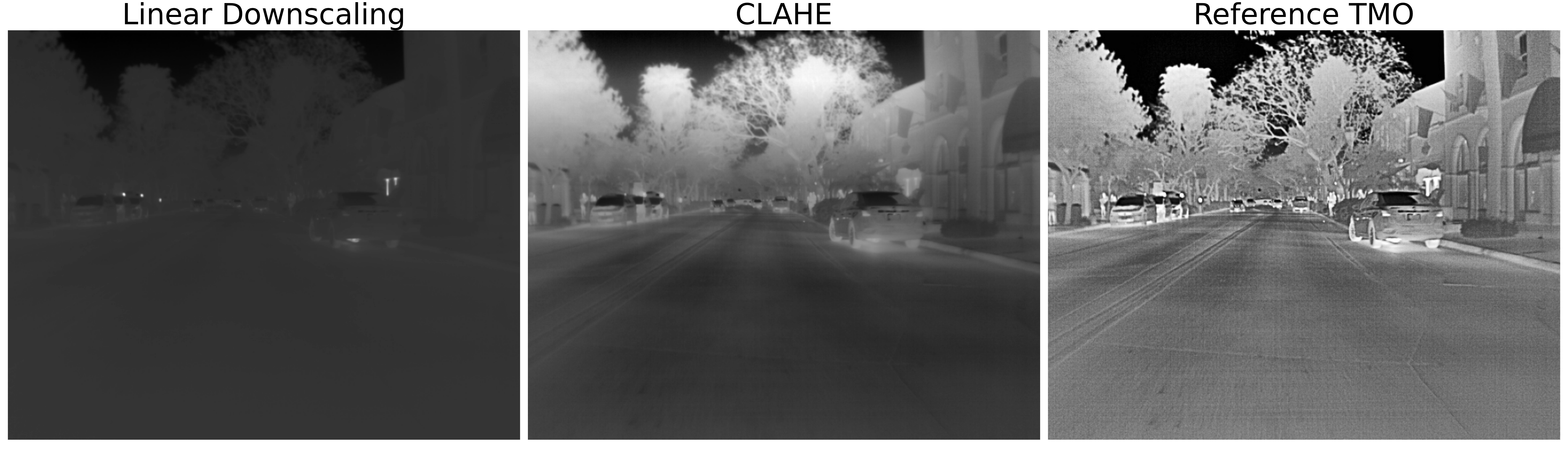}
\caption{Comparison between na{\"i}ve linear downscaling, CLAHE, and the optimized multi-scale Retinex-based reference TMO used in this paper. The example image is taken from the public FLIR ADAS Dataset~\cite{FLIR_dataset}.}
\label{fig:Comparison_downscaling}
\end{figure}

Scenes in the real world can exhibit a wide range of luminance values, far greater than what can be captured with any photographic sensor. Cameras digitize such scenes as pixel intensity values with a limited value range given by the available \ac{bpp}. \ac{HDR} cameras are able to capture those luminance values in a higher resolution through an increase in the number of \ac{bpp}. This allows for a finer discretization of those values and therefore a higher level of detail. This is especially important in \ac{TIR} imaging, where only one intensity channel is available and therefore a high number of \ac{bpp} is needed to maximize the contrast~\cite{Teutsch2020}. Showing this content on standard displays, however, comes with its own challenges as most monitors are limited to only 8\,\ac{bpp}. Figure~\ref{fig:Comparison_downscaling} shows some example images. Na{\"i}ve \ac{HDR} compression methods such as linear downscaling lead to a loss of the rich information contained in those \ac{HDR} images. Even popular image contrast enhancement methods such as \ac{CLAHE}~\cite{Zuiderveld1994} do not provide satisfying results. A \ac{TMO} aims at preserving the rich information while producing visually appealing images or videos at the same time~\cite{Eilertsen2013,Eilertsen2017}. In this paper, such \acp{TMO} are investigated for the compression of \ac{TIR} images and videos with 16\,\ac{bpp}. There are actually very few methods available at this time specifically dealing with tone mapping in the \ac{TIR} spectrum~\cite{Teutsch2020,Haibo_conditional_filtering_MSR}. Inspired by Luo et al.~\cite{Haibo_conditional_filtering_MSR}, the \ac{MSR} algorithm, originally intended for contrast enhancement~\cite{Petro2014}, is implemented and optimized as a reference for this purpose (see Fig.~\ref{fig:Comparison_downscaling}). This optimized \ac{MSR} algorithm is subsequently approximated or mimicked with a deep learning approach based on the popular U-Net architecture~\cite{Ronneberger_U-Net} in a fully supervised training scheme. Such a mimicking approach can not only save processing time due to the approximation~\cite{Chen_2017_ICCV}, but we will also use it to efficiently inject self-supervised denoising for the joint learning of tone mapping and image enhancement. The process of tone mapping often amplifies noise to such an extent that it becomes visible again~\cite{Eilertsen2017}. Therefore, either noise reduction during tone mapping, or active denoising afterwards is important to produce high quality images. We propose to reduce the remaining noise in the images implicitly by utilizing a self-supervised deep learning approach~\cite{Xu_2020_Noisy_as_clean} that can be jointly trained with the tone mapping approach. Performing joint denoising and tone mapping is not entirely new~\cite{Hu2022,Granados2015}, but our method is designed for \ac{TIR} images, uses most recent deep learning methods, and does not need full supervision due to the self-supervised denoising. Extensive experiments on the public FLIR ADAS Dataset\footnote{We use FLIR ADAS Dataset Version 1.3 August 16, 2019 (please see: https://adas-dataset-v2.flirconservator.com/).} provide a comprehensive comparison with the state-of-the-art and prove the effectiveness of our proposed method. Our contributions can be summarized as (1) an optimization of the \ac{MSR} algorithm to generate a reference \ac{TMO} for \ac{TIR} images and videos, (2) an approximation of this reference \ac{TMO} with a deep learning-based U-Net neural network architecture in a supervised training scheme, (3) the injection of self-supervised denoising for joint multi-task learning of tone mapping and denoising within the U-Net architecture, and (4) a deep discussion on self-supervised denoising and deflickering. Each part is analyzed in detail and accompanied by comprehensive experiments and results on the public FLIR ADAS Dataset~\cite{FLIR_dataset}.

The remainder of this paper is organized as follows: related work is reviewed in Section~\ref{sec:relwork}, the optimized \ac{MSR} algorithm based on traditional computer vision is introduced and evaluated in Section~\ref{sec:retinex}, deep learning-based tone mapping with joint implicit denoising is discussed in Section~\ref{sec:TMDCNN} together with extensive experimental results. The comparison with the state-of-the-art is provided in Section~\ref{sec:sota}. Conclusions are given in Section~\ref{sec:conclu}.

\section{RELATED WORK}
\label{sec:relwork}

\noindent\textbf{Tone mapping in the \ac{TIR} spectrum:} many methods for tone-mapping have been proposed over the last three decades~\cite{Devlin2002,Reinhard2010,Banterle:2017,Eilertsen2017}, but for the purpose of processing conventional color \ac{VIS} images. Approaches based on typical color conversions or color restoring processes are not considered here as \ac{TIR} images usually just have one intensity channel. However, since the brightness channel of an image is typically subjected to tone mapping, approaches can be adopted from the \ac{VIS} spectrum. Realtime TMO~\cite{Eilertsen2015} is a tone mapping algorithm based on traditional computer vision. It is still a state-of-the-art \ac{TMO} in terms of overall image quality in a comparative survey of different \acp{TMO}~\cite{Eilertsen2017}. It was utilized for tone mapping of \ac{TIR} images achieving good results and proving its suitability for this task~\cite{Teutsch2020}. \ac{MSR}~\cite{Petro2014} is another method taken from the \ac{VIS} spectrum that is based on traditional computer vision. It is a derivation from the Retinex theory proposed by Land and McCann \cite{Land1971}, which aimed to model how the \ac{HVS} perceives color in a scene. They established that the \ac{HVS} perceives brightness in a relative way, namely as variations of brightness in local image regions and not as absolute values.
The basic idea behind this algorithm is that an image can be split into its reflectance and illumination. The illumination value of a pixel $x$ is approximated as the ratio between the value of $x$ and the average value of the surrounding pixels. The reflectance is then calculated as the difference between the illumination image and the original image. Hyunchan et al. \cite{Hyunchan_Retinex_TMO} used a modified version of the \ac{MSR} algorithm for tone mapping of \ac{HDR} \ac{VIS} images. Luo et al. \cite{Haibo_conditional_filtering_MSR} successfully extended the \ac{MSR} approach with a \ac{CGF} for tone mapping of \ac{TIR} images, proving the suitability for this task. Further methods~\cite{Liu2019a,Liu2019b,Wan2018} are mostly based on contrast enhancement techniques, which is considered less promising for tone mapping~\cite{Teutsch2020}.

\noindent\textbf{Deep learning-based tone mapping:} traditional computer vision methods are still popular for tone mapping \ac{VIS} images and videos~\cite{Eilertsen2017, Shan_linear_window_TMO, Liang_L1L0_TMO, GU_Multiscale_Decomposition_TMO, Eilertsen2015,Haibo_conditional_filtering_MSR}. In recent years, however, more and more efforts have been made to approach this task with deep learning-based methods~\cite{GAN_TMO_Montulet2019DeepLF, GAN_TMO_Rana_2020, Patel_GAN_TMO, Su_U-Net_Deep_TMO, gharbi2017deep, Chen_2017_ICCV}. The main advantage is often seen in the efficiency gain that can theoretically be achieved by approximating these traditional \acp{TMO} with \acp{DCNN}~\cite{Chen_2017_ICCV}. A major problem of many tone mapping algorithms is their computational cost. Despite efforts to reduce the computation time, many of the current \acp{TMO} are not real-time capable even on high-performance hardware. Gharbi et al.~\cite{gharbi2017deep} address this problem and propose a combination of traditional and deep learning-based approaches to enable real-time tone mapping and general image enhancement on smartphones, achieving a frame rate of 40-50\,\ac{fps} on images with a resolution of $1920\times1080$\,pixels. Further potential for deep learning in the field of tone mapping also exists in possible quality improvements of the resulting images. Patel et al.~\cite{Patel_GAN_TMO} used a \ac{GAN}-based training strategy to train a mixture of multiple \acp{TMO} and obtained a higher \ac{TMQI} score for many of the tested images compared to the reference methods themselves. Furthermore, the results of \acp{DCNN} can potentially be further improved with different loss functions and methods of self-supervised learning. While general tone mapping for \ac{TIR} images is still a niche topic~\cite{Haibo_conditional_filtering_MSR}, deep learning-based tone mapping for \ac{TIR} images is almost non existent in the literature. Teutsch et al.~\cite{Teutsch2020} successfully trained the \ac{CAN} architecture, based on the works of Chen et al.~\cite{Chen_2017_ICCV}, to approximate the Realtime TMO~\cite{Eilertsen2015} in a supervised manner.

\noindent\textbf{Self-supervised denoising:} state-of-the-art denoising methods are based on deep learning nowadays and generally outperform denoising methods based on traditional computer vision. A majority of those learning-based denoising algorithms \cite{Vemulapalli_2016_CVPR_supervised_denoising, Zhang_supervised_denoising_2017, Zhang_supervised_denoising_2018} use supervised learning with clean data as \ac{GT}. Training data consists of noise-free, often synthetic, images that are corrupted with artificial noise. The \ac{ML} model is given the noisy images as input, and the corresponding clean images as target. 
In reality, however, the noise characteristics usually differ from the artificial noise models and domain gaps arise between synthetic and real images.
Recently, self-supervised approaches for image denoising gain increasing popularity. Noise2Noise~\cite{noise2noise-lehtinen} is one of the most important publications in this field. The approach is based on the assumption that noise is randomly distributed within an image and varies between consecutive frames of the same scene. Then, the noise of an infinite number of images averages out to zero over time.
Instead of training on clean images, they train on aligned, consecutive noisy image pairs of the same scene. Similar performance compared to fully supervised approaches can be achieved.
Since the existence of pixel-precise image alignment is a strong assumption, Ehret et al. \cite{Ehret_2019_CVPR} address this limitation and transfer this principle to the field of video denoising. Within a coherent video there are usually already several images of the same scene, which are, however, separated through time. To overcome this separation in time, Ehret et al. \cite{Ehret_2019_CVPR} used the optical flow of a pair of concurrent frames, to warp the last frame back to the current frame and thus remove the motion. Theoretically, this results in a pair of images that represent the same scene, but with different noise patterns. A denoising model can then be trained, based on the noise2noise \cite{noise2noise-lehtinen} approach. Ehret et al. \cite{Ehret_2019_CVPR} call this approach frame-to-frame training. A limiting factor of this method is the dependence on precise dense optical flow.
More recent deep learning-based denoising approaches have been focused on training \acp{DCNN} using only noisy images and without pairwise correspondence. Noisier2Noise~\cite{Moran_2020_CVPR_Noisier2Noise} and Noisy-As-Clean~\cite{Xu_2020_Noisy_as_clean} add additional noise to the original noisy image. This image pair can then be used for training a \ac{DCNN}. The Noisier2Noise approach works as follows. Let the natural noisy image $Y$ be a combination of the clean image without noise $X$ and the natural noise that should be removed $N$. Since only $Y$ is known, the goal is to get a prediction of $N$. For this, synthetic noise $M$ is added to the noisy image $Y$ to form a noisier version $Z$. $Z$ is then given as input to a \ac{DCNN}, while $Y$ is given as reference. The \ac{DCNN} then trains to map $Z$ to $Y$. Since the \ac{DCNN} only sees the input image as a whole, and therefore cannot distinguish between $N$ and $M$, the produced output is intuitively not perfect and somewhere between the input and the reference image. The predicted output is then substracted from the input $Z$ to gain an estimate of the noise which is still remaining in the prediction. This noise estimate is then subtracted from the prediction to gain the final denoised image. Similar to this approach, another approach was proposed by Xu et al.~\cite{Xu_2020_Noisy_as_clean} called Noisy-As-Clean. Just like in the Noisier2Noise approach, the natural noisy image $Y$ is given as reference and a with synthetic noise corrupted version $Z$ is given as input to a \ac{DCNN}. During training, the \ac{DCNN} learns to map $Y$ from $Z$. This version, however, differs from the Noisier2Noise approach in how inference is handled. For inference the trained model is given the natural noisy images $Y$ as input, but without the added noise and the output generated through this model is seen as the final denoised image.
The Noisier2Noise approach and the Noisy-As-Clean method have in common that they are trained in a completely self-supervised manner and that they can be trained directly on single images. No aligned image pairs or optical flow methods are needed.
Several of the mentioned self-supervised approaches are evaluated and compared on \ac{TIR} images and videos in this paper.

\noindent\textbf{Deflickering:} Flickering can be an issue when training a \ac{DCNN} on single images and then applying it to video data~\cite{Eilertsen2019}.
Under the assumption that there are no drastic light changes in the scene, consecutive images should have roughly similar intensity values. Zhang et al.~\cite{Zhang_2021_CVPR_Temporal_low_loght_video} used this principle to construct a loss function and enforce temporal consistency for the task of low-light image enhancement. Lai et al.~\cite{Lai_2018_ECCV_Temporal_deflickering_network} used the same principle to train a \ac{DCNN} for the task of deflickering.

\noindent\textbf{Multi-task learning:} the core idea of multi-task learning is that computer vision tasks exist, which are sufficiently similar that they can be learned jointly~\cite{zamir2018taskonomy}. In this way, each individual task can be further boosted in its performance since for example deep learning-based feature extraction in Mask R-CNN~\cite{He2017} benefits from the joint learning of object detection and instance segmentation. Joint learning of image enhancement tasks such as super resolution, low-light enhancement, dehazing, denoising, or tone mapping~\cite{Chen_2017_ICCV,Hu2022,Ma2022,Lyu2022,Xing2021} is obviously promising. Although it can be expected that multi-task learning can improve \ac{TIR} image enhancement as well, only very few literature exists up to now~\cite{KUANG2019119}. In this paper, we tackle joint learning of tone mapping, denoising, and deflickering for \ac{TIR} image and video enhancement.

\section{TONE MAPPING VIA MULTI-SCALE RETINEX}
\label{sec:retinex}

To establish a benchmark for the following experiments, we use an already existing evaluation framework for \ac{TIR} image and video tone mapping~\cite{Teutsch2020}, that we recently extended with a Python implementation of the \ac{TMQI}\footnote{https://github.com/HensoldtOptronicsCV/ToneMappingIQA}. Furthermore, we use the publicly available FLIR ADAS Dataset~\cite{FLIR_dataset}. We quantitatively assess the performance of the baseline \ac{TMO} approaches Realtime \ac{TMO}~\cite{Eilertsen2015}, \ac{CGF}~\cite{Haibo_conditional_filtering_MSR}, \ac{MSR}~\cite{Petro2014} and the original FLIR \ac{TMO} that was used to produce the 8~bit images of the FLIR dataset. As no public code was available for \ac{CGF}, we re-implemented the approach and use our code throughout this paper. The Realtime TMO implementation was adapted to the \ac{TIR} spectrum: it now processes single-channel \ac{TIR} images instead of three-channel color \ac{VIS} images and the parameterization was adjusted. Furthermore, the code of the FLIR \ac{TMO} is not publicly available and to the best of your knowledge the algorithm is not described in a paper. The dataset, however contains the original 16 bit and the tone mapped 8 bit images, so we can use them for the evaluation even without knowing the algorithm that produced the 8 bit images. 

These \acp{TMO} are evaluated on the dataset's \emph{video} subset. The results are shown in Table~\ref{tab:current_baseline}. The noise visibility measure needs a second set of images tone mapped on a noisy version of the original \ac{HDR} images. Since the FLIR \ac{TMO} is not publicly available, this set cannot be created and thus the noise visibility measure cannot be applied to the 8~bit images of the FLIR ADAS Dataset. 
The best results for each measure are written in bold. Recognizably, the 8~bit FLIR images perform best across most of the measures from the evaluation framework.
The \ac{CGF} performs better for the global temporal incoherence measure and for the overexposure measure, but produces the worst result for the underexposure measure, which indicates an imbalance of the pixel values towards the lower value range. When comparing the \ac{MSR} with the Realtime TMO, we can see that both perform very similar with the baseline \ac{MSR} being slightly better on the noise visibility measure and the \ac{TMQI}.

\begin{table}[ht]
\caption{Quantitative evaluation comparing different TMOs to form the current baseline.}
\begin{tabular}{rcccc}
    \multicolumn{1}{c}{\textbf{}}                    & \textbf{FLIR~\cite{FLIR_dataset}}                          & \textbf{Realtime TMO~\cite{Eilertsen2015}}          & \textbf{\ac{CGF}~\cite{Haibo_conditional_filtering_MSR}}   & \textbf{\ac{MSR}~\cite{Petro2014}} \\ \cline{2-5} 
    \multicolumn{1}{r|}{TMQI $\uparrow$}                        & \multicolumn{1}{c|}{\textbf{0.951}}   & \multicolumn{1}{c|}{0.809}  & \multicolumn{1}{c|}{0.850}   & \multicolumn{1}{c|}{0.866}\\ \cline{2-5} 
    \multicolumn{1}{r|}{Underexposure $\downarrow$}               & \multicolumn{1}{c|}{\textbf{0.535}}    & \multicolumn{1}{c|}{0.755}  & \multicolumn{1}{c|}{2.842}   & \multicolumn{1}{c|}{1.191}    \\ \cline{2-5} 
    \multicolumn{1}{r|}{Overexposure $\downarrow$}                & \multicolumn{1}{c|}{0.899}    & \multicolumn{1}{c|}{1.628}  & \multicolumn{1}{c|}{\textbf{0.030}}  & \multicolumn{1}{c|}{1.578}   \\ \cline{2-5} 
    \multicolumn{1}{r|}{Loss of Global Contrast $\downarrow$}     & \multicolumn{1}{c|}{\textbf{-0.162}}   & \multicolumn{1}{c|}{-0.123} & \multicolumn{1}{c|}{-0.084}  & \multicolumn{1}{c|}{-0.148}  \\ \cline{2-5} 
    \multicolumn{1}{r|}{Loss of Local Contrast $\downarrow$}      & \multicolumn{1}{c|}{\textbf{-0.0417}} & \multicolumn{1}{c|}{-0.0236} & \multicolumn{1}{c|}{-0.022}  & \multicolumn{1}{c|}{-0.027}  \\ \cline{2-5} 
    \multicolumn{1}{r|}{Noise Visibility $\downarrow$}            & \multicolumn{1}{c|}{-}                 & \multicolumn{1}{c|}{30.77}   & \multicolumn{1}{c|}{30.04}  & \multicolumn{1}{c|}{\textbf{28.79}}  \\ \cline{2-5} 
    \multicolumn{1}{r|}{Global Temporal Incoherence $\downarrow$} & \multicolumn{1}{c|}{$1.7\text{e}^{-4}$}  & \multicolumn{1}{c|}{$3.7\text{e}^{-4}$} &
    \multicolumn{1}{c|}{$\mathbf{2.8e^{-5}}$} &
    \multicolumn{1}{c|}{$2.7\text{e}^{-4}$} \\ \cline{2-5} 
    \multicolumn{1}{r|}{Local Temporal Incoherence $\downarrow$}  & \multicolumn{1}{c|}{\textbf{0.0048}}  & \multicolumn{1}{c|}{0.0157} & \multicolumn{1}{c|}{0.0088}    & \multicolumn{1}{c|}{0.0394} \\ \cline{2-5} 

\end{tabular}
\label{tab:current_baseline}
\end{table}

\begin{figure}[htbp]
\centering
\includegraphics[width=1\textwidth]{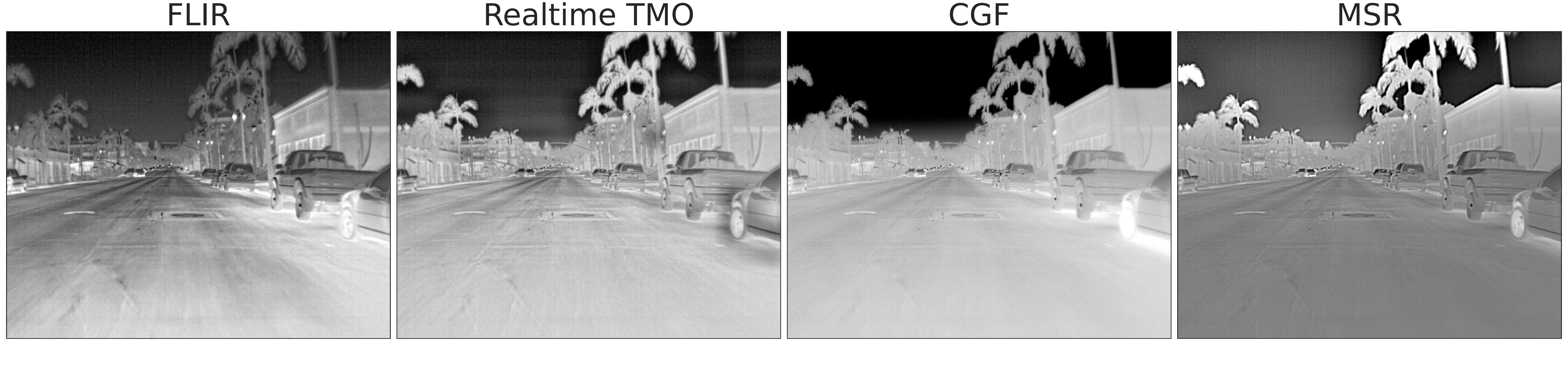}
\caption{Qualitative comparison between the different TMOs providing the current baseline.}
\label{fig:Comparison_baseline}
\end{figure}

A qualitative comparison of the \acp{TMO} is shown in Figure \ref{fig:Comparison_baseline}. The 8~bit FLIR images are visually appealing with high contrast and a good utilization of the 8~bit value range. The results of the Realtime TMO and especially those of the baseline \ac{MSR} algorithm have a weaker contrast. Comparing the results of the 8~bit FLIR images with those of \ac{CGF} and baseline \ac{MSR} reveals that Retinex-based algorithms produce a rather noisy, but also sharper images. To draw a conclusion, the qualitative evaluation roughly confirms the results of the quantitative evaluation that uses the publicly available evaluation framework.

While generally performing worse than the 8~bit FLIR images, the \ac{MSR} algorithm shows potential especially in terms of image sharpness. The main weaknesses of the algorithm at this stage are the low contrast, the amplification of image noise, and flickering artifacts that are noticeable in a video stream only. Those weaknesses, however, can potentially be addressed. Therefore, the next sections aim at optimizing the baseline \ac{MSR} algorithm.

\subsection{Image Contrast Enhancement}
\label{Retinex contrast enhancement}
One of the most obvious weaknesses of the \ac{MSR} algorithm is its low contrast and monotonous appearance. Very warm objects with a bright appearance in the image can be spotted easily. Details of darker objects or objects with similar brightness are hardly visible due to the weak contrast. To extract as much information as possible from an \ac{HDR} image, the subtle intensity differences in these areas should be emphasized. Therefore, the original \ac{MSR} algorithm~\cite{Petro2014} comes with a variant called Multi-Scale Retinex with Chromaticity Preservation (MSRCP). The idea is to perform an algorithm named \ac{SCB}~\cite{Limare2011} to use the full intensity range of each color channel. Since we use single-channel \ac{TIR} images in this paper, the proposed \ac{SCB} algorithm has turned out to be not really useful. Instead, one possible method of emphasizing the image details is \ac{HE}~\cite{gonzalez2008digital}. \ac{HE} aims at spreading the distribution of intensity values within the image histogram. This usually increases global contrast and thus produces a more visually appealing image. This approach, however, works globally on the entire image and therefore often lacks in enhancing the contrast of local image regions. To further increase local contrast, \ac{AHE}~\cite{gonzalez2008digital} can be used. This method performs tiling, i.e. it divides the input image into several regions. \ac{HE} is then performed on each region independently. As a downside, this method tends to increase image noise. The \ac{CLAHE} algorithm~\cite{Zuiderveld1994} was developed to mitigate this problem. This algorithm limits the contrast amplification after a predetermined threshold is reached.

\begin{figure}[htbp]
\centering
\includegraphics[width=1\textwidth]{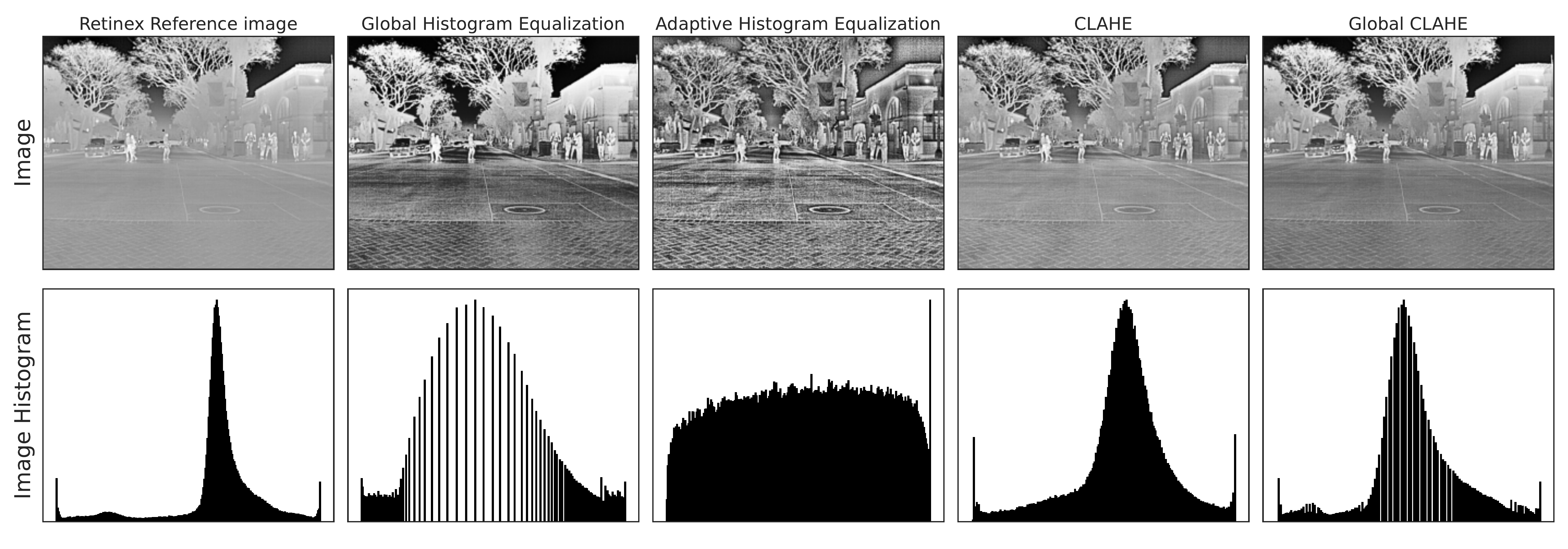}
\caption{Comparison between different methods for histogram equalization. The processed images are shown in the upper row and the related histograms in the lower row.}
\label{fig:comparison_histogram_equalization}
\end{figure}

All just described methods are implemented as a post-processing step after tone mapping. An alternative implementation as a pre-processing step was also tested, but lead to no noticeable differences in the output images. Figure~\ref{fig:comparison_histogram_equalization} shows an example image taken from the FLIR ADAS Dataset processed with each method together with its related histogram. The first image on the left is from the FLIR dataset tone mapped with the baseline \ac{MSR} algorithm. It serves as a reference. In the related histogram below the image, it can be seen that most of the gray-scale values are in a narrow range indicating weak contrast. Applying global \ac{HE} to this image visibly enhances the contrast, but also produces some gaps in the histogram, which means that the value range is again not well exploited.
\ac{AHE} manages to spread the values more evenly, but the global differences are less noticeable and the noise (mostly visible in the sky) is strongly amplified. In this example, we intentionally choose a rather small tile size to further highlight the differences between the considered methods. Limiting the contrast with the before mentioned \ac{CLAHE} algorithm (same tile size as \ac{AHE}) results in a less aggressive spread in the histogram and an overall more appealing image. Using a global implementation of \ac{CLAHE} without tiling does not improve the local contrast as much, but introduces less noise and gives the subjectively most visually appealing results. The global version of the \ac{CLAHE} algorithm will therefore be used as a post-processing step after the \ac{MSR} algorithm. This results in a clear improvement of contrast, while still preserving a natural appearance.

\subsection{Deflickering}
\label{Retinex deflickering}
One weakness of the \ac{MSR} algorithm is that it tends to introduce flickering. Rapid changes of the average image intensity between consecutive video frames are noticeable and disturbing. Such undesired global intensity changes are described as flickering in the literature~\cite{Guthier2011}.
\acp{TMO} designed for single image processing are likely to introduce flickering artifacts when they are applied to videos.

\begin{figure}[htbp]
\centering
\includegraphics[width=1\textwidth]{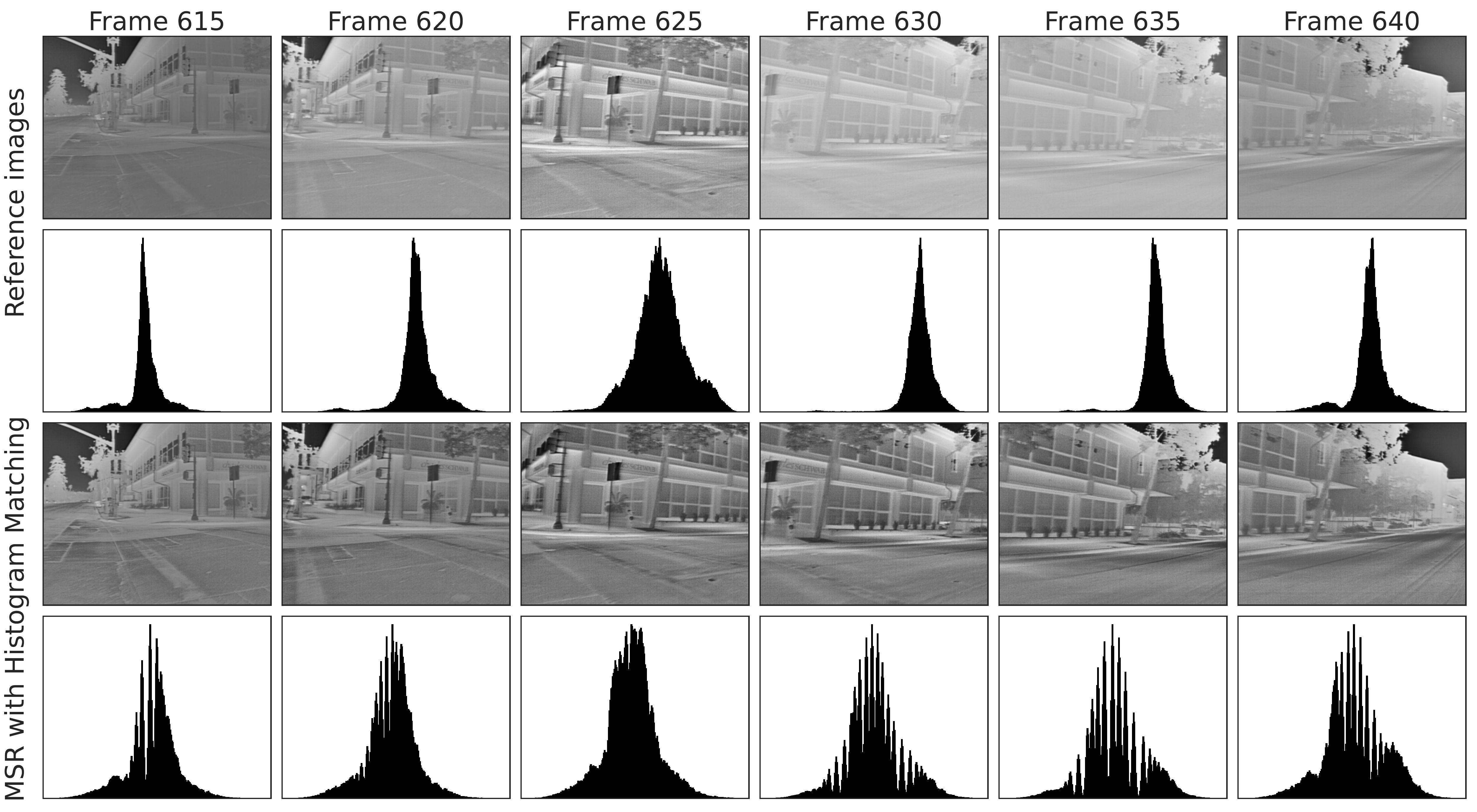}
 \caption{Comparison between the \ac{MSR} algorithm with and without deflickering through histogram matching. The example images are taken from the FLIR ADAS Dataset \emph{video} subset.}
\label{fig:comparison_deflickering}
\end{figure}

There are several ways to reduce flickering. Two different approaches are investigated is this paper. The first approach is based on histogram matching, which is a quite popular method for deflickering~\cite{Delon2006,Pitie2004,Kanj2017}. The goal of this approach is to transform the intensity values of an image so that its histogram after the transformation matches with a reference histogram~\cite{gonzalez2008digital}.
We calculate the average histogram of the 100 previous video frames as reference histogram. The histogram of the current frame is then matched to this reference histogram and the image intensity values are transformed using a lookup table.
The results are shown in Figure~\ref{fig:comparison_deflickering}. The frames for the comparison are intentionally chosen from a scene in the FLIR ADAS Dataset with rapidly changing global image intensity. To better illustrate the differences, only every fifth frame is shown. In the upper row are the reference frames tone mapped with the contrast enhanced \ac{MSR} approach and their related histograms are shown in the second row. In the third row are the results of the proposed deflickering method through histogram matching and their related histograms are shown in the bottom row. It can be observed that the rapid changes of global image intensity are smoothed over time while still maintaining a natural appearance and without introducing any visible artifacts.

\begin{figure}[htbp]
\centering
\includegraphics[width=1\textwidth]{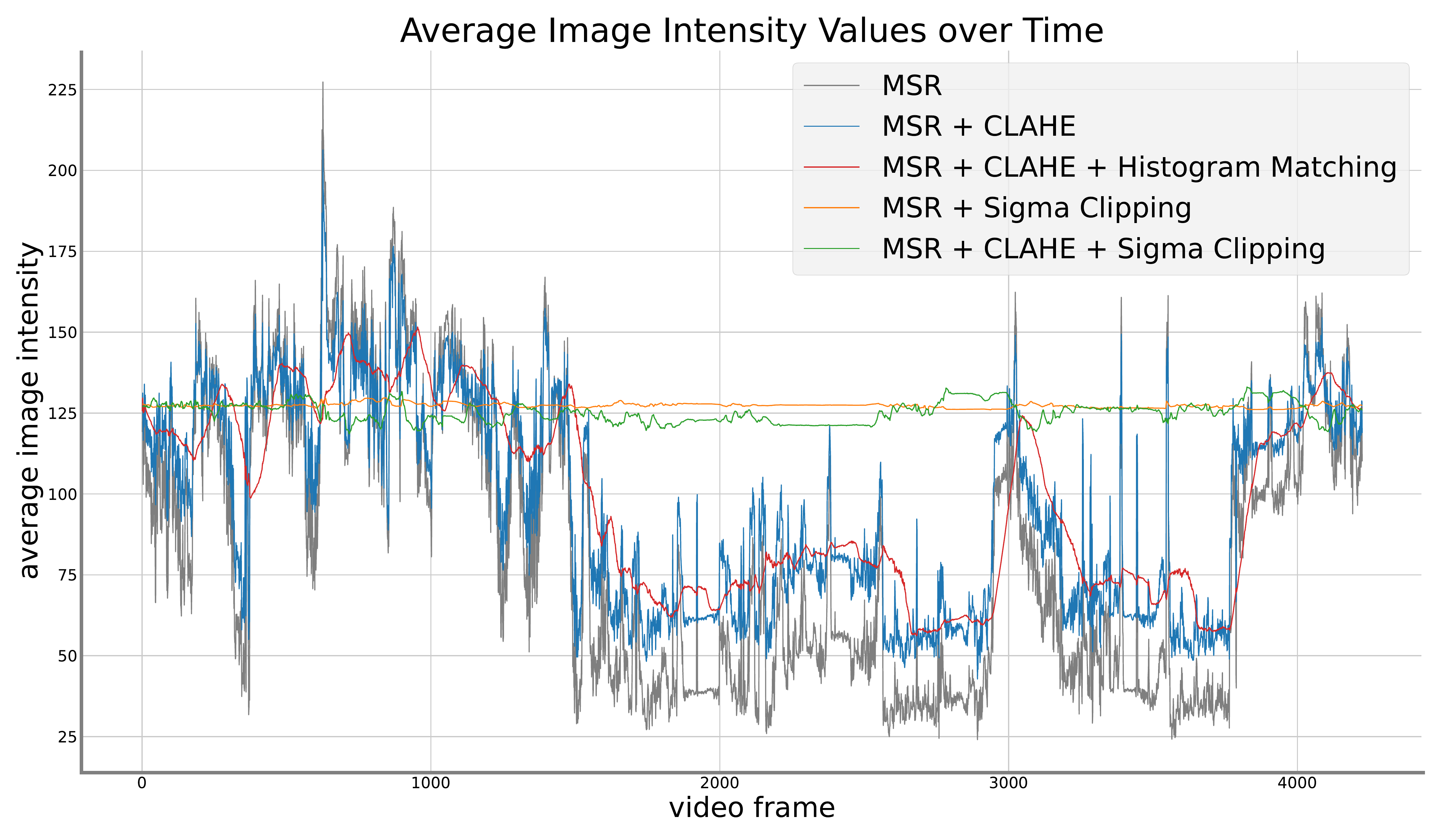}
\caption{Average image intensity value over time throughout the FLIR video dataset tone mapped with the \ac{MSR} algorithm with different combinations of CLAHE, sigma clipping and histogram matching. Both, sigma clipping and histogram matching smooth the average image intensity over time and thus perform deflickering.}
\label{fig:Mean_value_Sigma_vs_Simplest}
\end{figure}

Flickering can be detected by observing the average intensity value of an image over time. A high average value generally corresponds to a high global image intensity, i.e. a \emph{bright} image, and vice versa. A strong variation of this average intensity value between consecutive video frames indicates flickering. This effect is visualized in Figure~\ref{fig:Mean_value_Sigma_vs_Simplest}, where multiple different deflickering approaches are evaluated on the \emph{video} subset of the FLIR ADAS Dataset.
\ac{SCB} as proposed in the original \ac{MSR} paper~\cite{Petro2014} or CLAHE are not sufficient to compensate for flickering. Luo et al.~\cite{Haibo_conditional_filtering_MSR} instead performed clipping based on the standard deviation of the image. The mean value and the standard deviation are calculated for the image and intensity values that exceed three times the standard deviation get clipped. This very simple yet effective method, which also eliminates the need to manually choose a threshold for the clipping, is also called \emph{three-sigma rule} in the literature~\cite{Grafarend2006}. This method will be referred to as \emph{sigma clipping} in the remainder of this paper. In Figure~\ref{fig:Mean_value_Sigma_vs_Simplest}, we can see that sigma clipping also introduces some kind of equalization or normalization of the image by enforcing the average image intensity value staying close to 127. This is good for deflickering on the one hand, but may also be inflexible of capturing global scene temperature changes on the other hand. 
However, the FLIR ADAS Dataset does not provide real video data for the training of \ac{ML} approaches. Instead, frame skipped image sequences are provided in the \emph{training} subset. So, we cannot apply deflickering based on histogram matching when performing \ac{ML} in Section~\ref{sec:TMDCNN}. As a result, we will rather rely on sigma clipping when generating a \ac{MSR}-based \ac{GT} for approximating the reference \ac{TMO} with a deep neural network.

\subsection{Optimized Multi-Scale Retinex} \label{Optimized Retinex}
A quantitative evaluation of the optimization approaches analyzed in this section is given in Table~\ref{tab:Retinex_optimization}. Besides the state-of-the-art approaches FLIR, Realtime TMO, \ac{CGF}, and baseline \ac{MSR}, we also include contrast enhancement using global CLAHE as presented in Section~\ref{Retinex contrast enhancement} as well as sigma clipping as discussed in Section~\ref{Retinex deflickering}. Combining global \ac{CLAHE} for contrast enhancement and sigma clipping for deflickering forms the optimized version of the \ac{MSR} algorithm.
Both optimizations work well together and result in a notable improvement in image contrast, temporal coherence, and \ac{TMQI}. This improvement comes at the cost of stronger noise visibility. Therefore, different \ac{ML}-based approaches for noise reduction are investigated in the next section. The optimized \ac{MSR} is used as reference \ac{TMO} in the remainder of this paper.

\begin{table}[ht]
    \caption{Comparison between the state-of-the-art \acp{TMO} and the baseline \ac{MSR} algorithm with different optimizations. The proposed optimized \ac{MSR} performs best among the Retinex-based algorithms especially for the \ac{TMQI} measure.}
    \centering
    \resizebox{\textwidth}{!}{
    \begin{tabular}{rccccccc}
    \multicolumn{1}{c}{\textbf{}}                    & \textbf{FLIR~\cite{FLIR_dataset}}                       & \textbf{\begin{tabular}[c]{@{}c@{}}Realtime~\cite{Eilertsen2015}\\TMO\end{tabular}} & \textbf{\ac{CGF}~\cite{Haibo_conditional_filtering_MSR}} & \textbf{\ac{MSR}\cite{Petro2014}} & \textbf{\begin{tabular}[c]{@{}c@{}}\ac{MSR}+\\CLAHE\end{tabular}} & \textbf{\begin{tabular}[c]{@{}c@{}}\ac{MSR}+\\Sigma\\Clipping\end{tabular}} & \textbf{\begin{tabular}[c]{@{}c@{}}Optimized\\\ac{MSR}\end{tabular}} \\ \cline{2-8} 
    \multicolumn{1}{r|}{TMQI$\uparrow$}                        & \multicolumn{1}{c|}{\textbf{0.951}} & \multicolumn{1}{c|}{0.809}  & \multicolumn{1}{c|}{0.850}         & \multicolumn{1}{c|}{0.866}                                      & \multicolumn{1}{c|}{0.887}                                                                                         & \multicolumn{1}{c|}{0.921}           & \multicolumn{1}{c|}{0.923}                                           \\ \cline{2-8} 
    \multicolumn{1}{r|}{Underexposure$\downarrow$}               & \multicolumn{1}{c|}{\textbf{0.535}}          & \multicolumn{1}{c|}{0.755}  & \multicolumn{1}{c|}{2.842}   & \multicolumn{1}{c|}{1.191}                                      & \multicolumn{1}{c|}{1.212}                                                                         & \multicolumn{1}{c|}{1.685}                  & \multicolumn{1}{c|}{1.803}                         \\ \cline{2-8} 
    \multicolumn{1}{r|}{Overexposure$\downarrow$}                & \multicolumn{1}{c|}{0.899}          & \multicolumn{1}{c|}{1.628} & \multicolumn{1}{c|}{\textbf{0.030}}   & \multicolumn{1}{c|}{1.580}                                      & \multicolumn{1}{c|}{2.437}                                                                       & \multicolumn{1}{c|}{0.860}                     & \multicolumn{1}{c|}{1.176}                                \\ \cline{2-8} 
    \multicolumn{1}{r|}{Loss of Global Contrast$\downarrow$}     & \multicolumn{1}{c|}{-0.162}         & \multicolumn{1}{c|}{-0.123}   & \multicolumn{1}{c|}{-0.084}   & \multicolumn{1}{c|}{-0.148}                                     & \multicolumn{1}{c|}{\textbf{-0.204}}               & \multicolumn{1}{c|}{-0.151}                       & \multicolumn{1}{c|}{-0.202}                              \\ \cline{2-8} 
    \multicolumn{1}{r|}{Loss of Local Contrast$\downarrow$}      & \multicolumn{1}{c|}{-0.0417}        & \multicolumn{1}{c|}{-0.0233}  & \multicolumn{1}{c|}{-0.022}  & \multicolumn{1}{c|}{-0.0270}                                    & \multicolumn{1}{c|}{-0.0425}               & \multicolumn{1}{c|}{-0.0290}                   & \multicolumn{1}{c|}{\textbf{-0.0447}}                                 \\ \cline{2-8} 
    \multicolumn{1}{r|}{Noise Visibility$\downarrow$}            & \multicolumn{1}{c|}{-}              & \multicolumn{1}{c|}{30.77}  & \multicolumn{1}{c|}{30.04}    & \multicolumn{1}{c|}{\textbf{28.79}}                                      & \multicolumn{1}{c|}{30.61}                                     & \multicolumn{1}{c|}{29.50}                            & \multicolumn{1}{c|}{30.67}                          \\ \cline{2-8} 
    \multicolumn{1}{r|}{Global Temporal Incoherence$\downarrow$} & \multicolumn{1}{c|}{$1.7\text{e}^{-4}$}       & \multicolumn{1}{c|}{$3.6\text{e}^{-4}$}  & \multicolumn{1}{c|}{$2.8\text{e}^{-5}$}  & \multicolumn{1}{c|}{$2.7\text{e}^{-4}$}                                   & \multicolumn{1}{c|}{$1.5\text{e}^{-4}$}               & \multicolumn{1}{c|}{$\mathbf{1.6e^{-5}}$}                      & \multicolumn{1}{c|}{$2\text{e}^{-5}$}                              \\ \cline{2-8} 
    \multicolumn{1}{r|}{Local Temporal Incoherence$\downarrow$}  & \multicolumn{1}{c|}{\textbf{0.0048}}        & \multicolumn{1}{c|}{0.0158} & \multicolumn{1}{c|}{0.0088} & \multicolumn{1}{c|}{0.0394}                                     & \multicolumn{1}{c|}{0.0253}                                                                        & \multicolumn{1}{c|}{0.0389}                      & \multicolumn{1}{c|}{0.025}                               \\ \cline{2-8} 
    \end{tabular}
    }
    \label{tab:Retinex_optimization}
\end{table}

\section{DEEP LEARNING BASED TONE MAPPING}
\label{sec:TMDCNN} 
Our motivation to use deep learning for tone mapping is that we can efficiently inject further methods into the training algorithm that perform image enhancement such as denoising. Therefore, we first need to train a deep neural network to approximate or mimic the related traditional computer vision-based \ac{TMO}. We test to mimic both the optimized \ac{MSR} and the FLIR \ac{TMO} in two separate experiments. So, the 8~bit output images of the optimized \ac{MSR} algorithm and the FLIR \ac{TMO} serve as \ac{GT} during training. Hence, the basic supervised learning approach for tone mapping using \acp{DCNN} is straight forward. First, a set of \ac{HDR} images is tone mapped with a preferred \ac{TMO} that can be utilized as reference images or targets for the training process. The resolution of the \ac{DCNN}'s input layer is 640$\times$512~pixels, which is the original resolution of the FLIR ADAS Dataset. The same \ac{HDR} images are then propagated through the \ac{DCNN} to form an output. This output is compared with the respective reference image and the error, or difference between both images is calculated through a loss function and used to train the \ac{DCNN} via backpropagation. Just like other popular methods for deep learning-based image enhancement~\cite{Chen_2017_ICCV,Montulet_Briassouli_U-Net_Deep_TMO}, we use the \ac{MSE} loss.

We consider two different architectures for the \ac{DCNN}: U-Net~\cite{Ronneberger_U-Net} and \ac{CAN}~\cite{CAN_YU_Koltun}.
The U-Net architecture was first proposed in 2015 for the application to medical image segmentation~\cite{Ronneberger_U-Net}. In a recent study~\cite{Komatsu_Gonsalves_U-Net_denoising}, different U-Net architectures where compared against each other and against other state-of-the-art models for the task of denoising. While the results of the standard U-Net architecture were at least comparable with the other models, the use of group normalization~\cite{Wu_2018_ECCV} increased the performance of the U-Net architecture, to consistently outperform other state-of-the-art denoising models. The U-Net architecture was also successfully used for tone mapping by Montulet and Briassouli \cite{Montulet_Briassouli_U-Net_Deep_TMO} who paired this architecture with a \ac{GAN} based training approach. For the experiments, we implemented the U-Net architecture in PyTorch based on the original architecture~\cite{Ronneberger_U-Net}. The main difference between the original and our implementation lies in the input and output layers, which were modified for the task of tone mapping with grayscale images. Other than that, batch normalization layers and later group normalization layers were added to improve the overall performance. The model width (number of feature maps in the first layer) and depth (number of consecutive layers) are tuned for approximating the optimized \ac{MSR} \ac{TMO} using automatic hyperparameter tuning. A depth of 6 and a with of 16 give us the best results for this particular task and are therefore used throughout the experiments.

The \ac{CAN} architecture was originally developed for semantic segmentation~\cite{CAN_YU_Koltun}. Chen et al.~\cite{Chen_2017_ICCV} proved the versatility of this architecture through successfully applying it to different image processing tasks, one of them being tone mapping for \ac{VIS} images. They also introduced an adaptive batch normalization step to the network, which further improved overall performance and the versatility of the architecture. However, in our implementation of the \ac{DCNN} architecture, this adaptive batch normalization caused stability issues during training. Instead, we used batch normalization at the beginning and later switched to group normalization, which completely eliminated padding artifacts that were visible with batch normalization. Padding artifacts appear as disturbing horizontal and vertical lines at the image borders. This seems to be the result of the dilated convolutions, which are extensively used in the \ac{DCNN} architecture. The \ac{CAN} architecture is implemented in PyTorch and it is based on the publicly available TensorFlow implementation~\cite{Chen_2017_ICCV}. The model depth (number of consecutive layers) and the model width (number of feature maps for each layer) are tuned for approximating the optimized \ac{MSR} \ac{TMO} using automatic hyperparameter tuning. We achieve the best results with a width of 24 and a depth of 7.

\begin{figure}[htbp]
\centering
\includegraphics[width=1\textwidth]{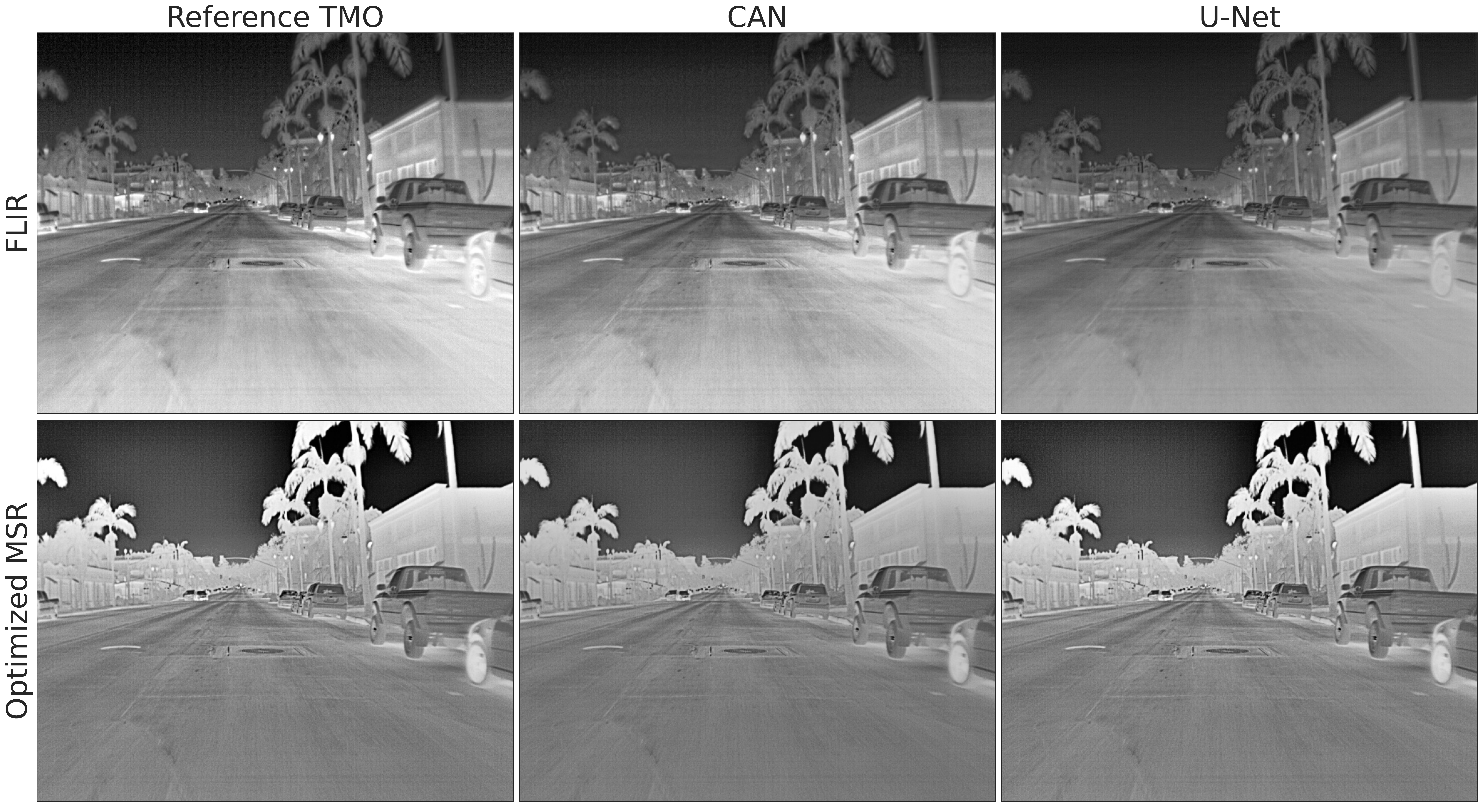}
\caption{Comparison between the results of the deep learning-based approximation. The idea is to mimic the performance of reference \acp{TMO} based on traditional computer vision by deep neural networks. The example images show that the two tested neural network architectures CAN and U-Net can successfully mimic the reference \acp{TMO}. Interestingly, CAN performs better when mimicking the FLIR \ac{TMO}, while U-Net performs better when mimicking the optimized \ac{MSR}.}
\label{fig:comparison_CAN_U-Net}
\end{figure}

Both the U-Net and \ac{CAN} models are trained for 32 epochs. As we aim at mimicking both the optimized \ac{MSR} algorithm and the FLIR \ac{TMO}, we use the 8~bit output images of the optimized \ac{MSR} and the original 8~bit images from the FLIR ADAS Dataset. The models did not show significant improvement with longer training and the use of group normalization led to consistent improvements for both architectures compared to batch normalization. In terms of qualitative results, shown in Figure~\ref{fig:comparison_CAN_U-Net}, the \ac{CAN} architecture mimic both original approaches very well, while U-Net struggled with the FLIR dataset and produced darker and less vibrant output images. However, U-Net performed slightly better than \ac{CAN} for most measurements of the optimized \ac{MSR}, except for noise visibility, as shown in Table~\ref{tab:Comparison_CAN_UNet}. Overall, both methods are suitable for approximating the optimized \ac{MSR} with \ac{CAN} performing better on the FLIR ADAS Dataset, and U-Net performing slightly better on the optimized \ac{MSR} \ac{TMO}. Since U-Net is also widely used for denoising, it is chosen as \ac{DCNN} architecture for the remainder of this paper.

\begin{table}[ht]
    \caption{Comparison between the FLIR \ac{TMO}, our optimized MSR and our deep learning-based approximation using the CAN and the U-Net architecture. Both methods can be approximated according to the measures. Mimicking the optimized \ac{MSR} works slightly better according to the \ac{TMQI}.}
    \centering
    \resizebox{\textwidth}{!}{
    \begin{tabular}{rcccccc}
    \multicolumn{1}{c}{\textbf{}}                    & \textbf{FLIR~\cite{FLIR_dataset}}    & \textbf{\begin{tabular}[c]{@{}c@{}}CAN\\FLIR\end{tabular}} & \textbf{\begin{tabular}[c]{@{}c@{}}U-Net\\FLIR\end{tabular}}    & \textbf{\begin{tabular}[c]{@{}c@{}}Optimized \\MSR\end{tabular}}   & \textbf{\begin{tabular}[c]{@{}c@{}}CAN\\Optimized\\MSR\end{tabular}}  & \textbf{\begin{tabular}[c]{@{}c@{}}U-Net\\Optimized\\MSR\end{tabular}}\\ \cline{2-7} 
    \multicolumn{1}{r|}{TMQI$\uparrow$}   & \multicolumn{1}{c|}{\textbf{0.951}} & \multicolumn{1}{c|}{0.939} & \multicolumn{1}{c|}{0.917} & \multicolumn{1}{c|}{0.923}    & \multicolumn{1}{c|}{0.926}  & \multicolumn{1}{c|}{0.933} \\ \cline{2-7} 
    \multicolumn{1}{r|}{Underexposure$\downarrow$}  & \multicolumn{1}{c|}{0.535}  & \multicolumn{1}{c|}{0.0112} & \multicolumn{1}{c|}{\textbf{0.0088}} & \multicolumn{1}{c|}{1.803}     & \multicolumn{1}{c|}{0.372}  & \multicolumn{1}{c|}{0.334} \\ \cline{2-7} 
    \multicolumn{1}{r|}{Overexposure$\downarrow$}   & \multicolumn{1}{c|}{0.899} & \multicolumn{1}{c|}{0.0375} & \multicolumn{1}{c|}{\textbf{0.0025}} & \multicolumn{1}{c|}{1.176}     & \multicolumn{1}{c|}{0.749}  & \multicolumn{1}{c|}{0.494} \\ \cline{2-7} 
    \multicolumn{1}{r|}{Loss of Global Contrast$\downarrow$}     & \multicolumn{1}{c|}{-0.162} & \multicolumn{1}{c|}{-0.100} & \multicolumn{1}{c|}{-0.0948} & \multicolumn{1}{c|}{\textbf{-0.202}}    & \multicolumn{1}{c|}{-0.168}   & \multicolumn{1}{c|}{-0.176} \\ \cline{2-7} 
    \multicolumn{1}{r|}{Loss of Local Contrast$\downarrow$}      & \multicolumn{1}{c|}{-0.0417} & \multicolumn{1}{c|}{-0.0425} & \multicolumn{1}{c|}{\textbf{-0.0482}} & \multicolumn{1}{c|}{-0.0447}   & \multicolumn{1}{c|}{-0.0444}   & \multicolumn{1}{c|}{-0.0474} \\ \cline{2-7} 
    \multicolumn{1}{r|}{Noise Visibility$\downarrow$}            & \multicolumn{1}{c|}{-}  & \multicolumn{1}{c|}{\textbf{27.66}} & \multicolumn{1}{c|}{27.68} & \multicolumn{1}{c|}{30.67}      & \multicolumn{1}{c|}{30.37}   & \multicolumn{1}{c|}{30.61} \\ \cline{2-7} 
    \multicolumn{1}{r|}{Global Temporal Incoherence$\downarrow$} & \multicolumn{1}{c|}{$1.7\text{e}^{-4}$}  & \multicolumn{1}{c|}{$1\text{e}^{-3}$} & \multicolumn{1}{c|}{$1.9\text{e}^{-3}$} & \multicolumn{1}{c|}{\textbf{$\mathbf{2e^{-5}}$}}     & \multicolumn{1}{c|}{$1.2\text{e}^{-3}$}  & \multicolumn{1}{c|}{$2.5\text{e}^{-4}$} \\ \cline{2-7} 
    \multicolumn{1}{r|}{Local Temporal Incoherence$\downarrow$}  & \multicolumn{1}{c|}{0.0048} & \multicolumn{1}{c|}{0.0014} & \multicolumn{1}{c|}{\textbf{0.0012}} & \multicolumn{1}{c|}{0.025}       & \multicolumn{1}{c|}{0.0191}   & \multicolumn{1}{c|}{0.0186} \\ \cline{2-7} 
    \end{tabular}
    }
    \label{tab:Comparison_CAN_UNet}
\end{table}
\subsection{Self-supervised Denoising}
\label{sec:denoise}
When considering the current results, further optimization potential can be found in the reduction of the image noise amplified by the Retinex algorithm.
The main goal is to find a denoising method that reduces the noise of the tone mapped images as much as possible, while preserving the image details at the same time. This method should also be compatible and easy to inject into the deep learning-based tone mapping approaches presented in the previous section.
For this purpose, three self-supervised denoising approaches are investigated, namely the model-blind frame-to-frame denoising approach by Ehret et al.~\cite{Ehret_2019_CVPR}, the Noisier2Noise approach by Moran et al.~\cite{Moran_2020_CVPR_Noisier2Noise}, and the Noisy-As-Clean approach by Xu et al.~\cite{Xu_2020_Noisy_as_clean}.

\begin{figure}[htbp]
\centering
\includegraphics[width=1\textwidth]{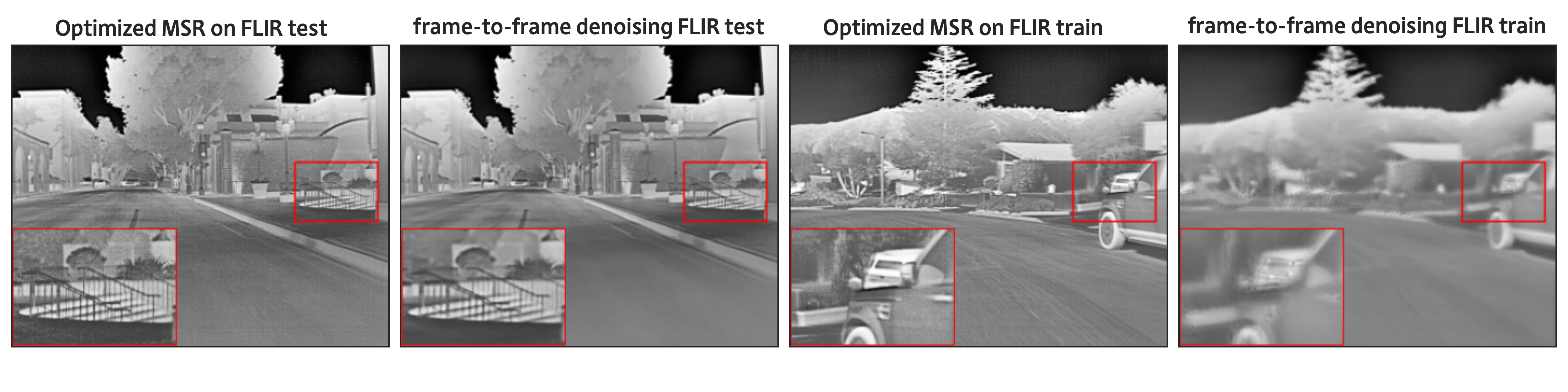}
\caption{Results of the frame-to-frame denoising approach on the FLIR test set and the FLIR train set. From left to right: reference image without denoising from FLIR test set, the same image after frame-to-frame denoising, reference image without denoising from FLIR train set, the same image after frame-to-frame denoising.}
\label{fig:Comparison_F2F_denoising}
\end{figure}

The frame-to-frame video denoising method proposed by Ehret et al. \cite{Ehret_2019_CVPR} uses dense optical flow to warp the previous frame to the current frame. This creates an image pair of the same scene, but with randomly distributed and thus independent noise. Denoising can then be trained through the original Noise2Noise approach~\cite{noise2noise-lehtinen}. The original code by Ehret et al. is publicly available. It is based on an online training technique, where training and inference are performed for each new frame. The results are depicted in Figure~\ref{fig:Comparison_F2F_denoising}. It can be clearly seen that the frame-to-frame approach struggles to handle images from the \emph{FLIR train} subset of the FLIR ADAS Dataset (images on the right). This is presumably due to the subset's low frame rate. The frame rate of about 2~\ac{fps} in combination with the fact that the camera is attached to a moving vehicle causes too much change in scene appearance between consecutive frames. Using a higher frame rate, like in the \emph{FLIR test} set (on the left) produces better results. While doing a good job in terms of denoising, it can be seen that the frame-to-frame approach produces blurry results on those images. This is most likely due to the fact that dense optical flow suffers from the lack of texture~\cite{Teutsch2021} in \ac{TIR} imagery. Ehret et al.~\cite{Ehret_2019_CVPR} use the TV-L1 method~\cite{Zach_TV-L1_optical_flow,Perez_TV-L1_optical_flow_impl} to calculate dense optical flow. We tested two popular alternative methods for calculating the dense optical flow~\cite{gunnar_farneback_flow,Yang_and_Ramanan_2019_VCN-Net_flow} but the results got even worse.
As a result, the calculation of dense optical flow is too imprecise on the FLIR ADAS Dataset and it is not possible to generate precisely aligned image pairs for the frame-to-frame training approach. It can be concluded that this video denoising approach is not suitable for our task here.

Both the Noisier2Noise denoising approach proposed by Moran et al.~\cite{Moran_2020_CVPR_Noisier2Noise}, and the Noisy-As-Clean denoising approach, proposed by Xu et al.~\cite{Xu_2020_Noisy_as_clean} use data augmentation of the input image combined with deep learning to denoise an input image. The core idea is to add artificially generated synthetic noise to the already noisy real training image and thus learn implicit denoising. 
However, while in the Noisier2Noise approach the synthetic noise is added to the input image for both training and inference, the Noisy-As-Clean approach uses the added synthetic noise only for the training phase.
Both approaches work on a single image basis, in contrast to the frame-to-frame approach. They are therefore well-suited for the low frame rate FLIR ADAS Dataset. The U-Net architecture~\cite{Ronneberger_U-Net} is used as network architecture for both approaches. For the augmentation of the input training images, additive Poisson noise with a lambda value of 100 is used, since this is expected to be closest to the real noise that is present on the FLIR dataset~\cite{Kennedy1993,Isoz2005}.

\begin{figure}[htbp]
\centering
\includegraphics[width=1\textwidth]{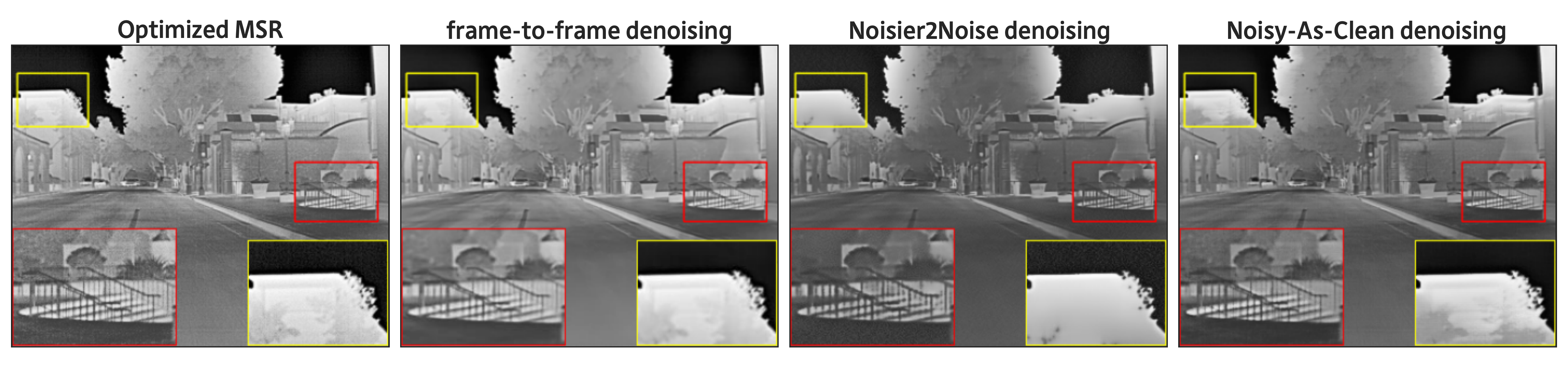}
\caption{Comparison between different self-supervised denoising approaches. While frame-to-frame denoising~\cite{Ehret_2019_CVPR} introduces image blur and Noisier2Noise~\cite{Moran_2020_CVPR_Noisier2Noise} loses image details, the Noisy-As-Clean~\cite{Xu_2020_Noisy_as_clean} approach provides the best denoising performance.}
\label{fig:Comparison_noisyasclean_denoising}
\end{figure}

The results are shown in Figure~\ref{fig:Comparison_noisyasclean_denoising}.
The frame-to-frame approach shows a notable reduction in noise, which is also confirmed by a quantitative analysis as the noise visibility is reduced from 30.67 to 26.93~dB. However, this method tends to blur image details as mentioned earlier. The Noisier2Noise approach produces inconsistent results across the image. While some areas, such as the railing in the middle right, display low levels of noise, others, such as the sky, are still noisy. Furthermore, certain image details are lost with this approach. This can be observed in the building in the top left of the image, where the details of the tree are completely washed out. This approach also reduces the global brightness and the contrast. The Noisy-As-Clean approach led to a consistent improvement in denoising in comparison to Noisier2Noise coming closer to that of the frame-to-frame approach, while also preserving image sharpness. The overall image details are also preserved together with the image contrast.
Overall, the Noisy-As-Clean approach is considered as the most promising for this work and is therefore further investigated. Quantitative results are depicted in Table~\ref{tab:Denoising_and_Deflickering}. 

\subsection{Joint Tone Mapping and Denoising}
\label{sec:jointTMOD}
\begin{figure}[htbp]
\centering
\includegraphics[width=1\textwidth]{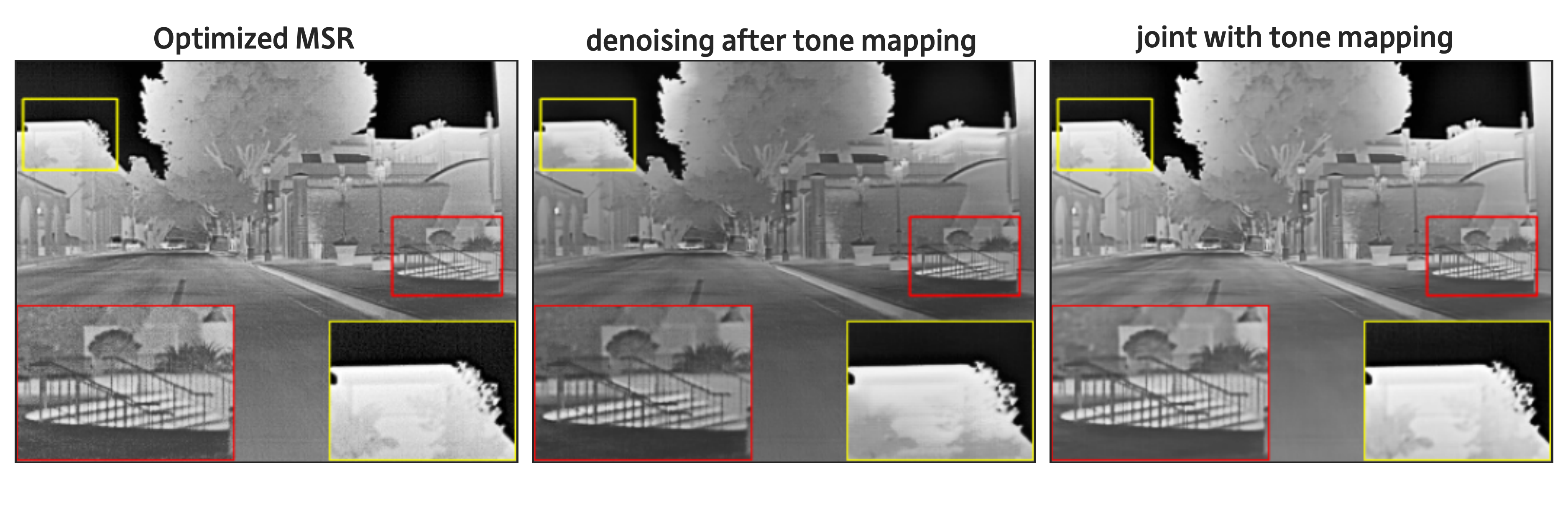}
\caption{Comparison between the self-supervised Noisy-As-Clean denoising approach~\cite{Xu_2020_Noisy_as_clean} used as post-processing, or learned jointly together with tone mapping.}
\label{fig:Comparison_joint_TM_denoising}
\end{figure}

Since the self-supervised denoising approach of our choice~\cite{Xu_2020_Noisy_as_clean} only uses augmentation of the input images during training, it can be injected easily into the training process that learns the approximation of the \acs{TMO}. For this, the U-Net architecture is used with the same configuration and in the same manner as for the supervised training of tone mapping. The only difference is that the input \ac{HDR} images are corrupted with additive Poisson noise with a lambda value of 100 during training. The results of this joint supervised learning of tone mapping and self-supervised denoising are shown in Figure~\ref{fig:Comparison_joint_TM_denoising}. We also added an image example, in which the same denoising approach was trained separately as a post-processing step after learning to approximate the \ac{TMO}. It can be seen that the denoising performance has even increased during joint learning, which can be confirmed by the image measures in Table~\ref{tab:Denoising_and_Deflickering}, while the overall appearance and contrast is closer to the original image. This suggests that training this approach together with the task of tone mapping enriches the denoising performance, while not negatively impacting the tone mapping. Training and performing both tasks at the same time in a single network also comes with performance benefits and a decrease of complexity.  

\subsection{Deflickering} \label{sec:deep_deflickering}
Deep learning-based approximation of the optimized \ac{MSR} \ac{TMO} introduces flickering again, which is even further increased when combined with denoising. This is not unexpected since the learning approach only considers single images during training.
A popular approach to enforce temporal consistency of consecutive video frames in deep learning is the use of dense optical flow~\cite{Zhang_2021_CVPR_Temporal_low_loght_video,Lai_2018_ECCV_Temporal_deflickering_network}.
Since the experiments with the frame-to-frame denoising approach already showed that calculating dense optical flow produces imprecise results for the FLIR ADAS Dataset, approaches based on optical flow calculation are not further investigated here. 
Instead, we perform a very simple approach introducing a temporal regularization loss term that penalizes changes in the average image intensity between consecutive video frames. In this way, temporal consistency is enforced.
The temporal regularization loss term $\mathcal{L}_\textit{reg}$ is calculated as:
\begin{equation}
\label{eq:TemporalLoss}
\mathcal{L}_\textit{reg}=\frac{1}{2}(\Bar{x}_{t}-\frac{1}{n}\sum_{k=1}^{n} \Bar{x}_{t-k})^2  
\end{equation}
 
The average image intensity value of the current output image $\Bar{x}_{t}$ is compared with the uniformly weighted sum of the average values of the last $n$ output images $\Bar{x}_{t-k}$. A high deviation between the current average intensity value and the past values leads to a high loss and is therefore penalized during training. In our experiments, we set $n=10$. The total loss $\mathcal{L}_\textit{total}$ is then calculated as a weighted sum of the standard \ac{MSE} loss term $\mathcal{L}_\textit{MSE}$ and the temporal regularization loss term:
\begin{equation}
\label{eq:TotalLoss}
\mathcal{L}_\textit{total}=\alpha\mathcal{L}_\textit{MSE} + (1-\alpha)\mathcal{L}_\textit{reg} 
\end{equation}

In our experiments, we set $\alpha$ to 0.9. The impact of this deflickering method on the global temporal incoherence measure is shown in Table~\ref{tab:Denoising_and_Deflickering}.
The remarkable reduction in global temporal incoherence is equivalent to a noticeable deflickering performance, which is confirmed by a subjective evaluation of the video stream. Also noticeable is an increase in the number of under- and overexposed pixels coming closer to the optimized \ac{MSR}. This can be explained by the reduction of outliers using the temporal regularization loss term. Outliers lead to a rather poor utilization of the 8bit value range and therefore to a reduction in the over- and underexposed pixels.

\begin{table}[ht]
    \caption{Quantitative evaluation of the self-supervised denoising method and denoising with deflickering. The considered denoising approach is first evaluated as post-processing after the approximation of tone mapping. Then, joint training is done with implicit denoising and deflickering. We use Noisy-As-Clean as denoising method in the joint training. Joint TMO+denoising+deflickering performs best according to the measures.}
    \centering
    \begin{tabular}{rcccc}
    \multicolumn{1}{c}{\textbf{}}                    & \textbf{\begin{tabular}[c]{@{}c@{}}Optimized\\MSR\end{tabular}}    & \textbf{\begin{tabular}[c]{@{}c@{}}Noisy-As-Clean\\denoising as\\post-processing\end{tabular}} & \textbf{\begin{tabular}[c]{@{}c@{}}Joint TMO\\+denoising\end{tabular}} & \textbf{\begin{tabular}[c]{@{}c@{}}Joint TMO\\+denoising\\+deflickering\end{tabular}} \\ \cline{2-5} 
    \multicolumn{1}{r|}{TMQI$\uparrow$}                        & \multicolumn{1}{c|}{0.923}  & \multicolumn{1}{c|}{\textbf{0.937}}                                                 & \multicolumn{1}{c|}{\textbf{0.937}}                                         & \multicolumn{1}{c|}{0.934}                                                      \\ \cline{2-5} 
    \multicolumn{1}{r|}{Underexposure$\downarrow$}               & \multicolumn{1}{c|}{1.803}                                                & \multicolumn{1}{c|}{\textbf{0.430}}                                               & \multicolumn{1}{c|}{0.460}                                            & \multicolumn{1}{c|}{1.330}                                           \\ \cline{2-5} 
    \multicolumn{1}{r|}{Overexposure$\downarrow$}                & \multicolumn{1}{c|}{1.176}                                              & \multicolumn{1}{c|}{0.439}                                     & \multicolumn{1}{c|}{\textbf{0.188}}                     & \multicolumn{1}{c|}{1.110}                                                     \\ \cline{2-5} 
    \multicolumn{1}{r|}{Loss of Global Contrast$\downarrow$}     & \multicolumn{1}{c|}{\textbf{-0.202}}        & \multicolumn{1}{c|}{-0.145}                                   & \multicolumn{1}{c|}{-0.133}                                 & \multicolumn{1}{c|}{-0.152}                                                     \\ \cline{2-5} 
    \multicolumn{1}{r|}{Loss of Local Contrast$\downarrow$}      & \multicolumn{1}{c|}{\textbf{-0.044}}        & \multicolumn{1}{c|}{-0.034}                           & \multicolumn{1}{c|}{-0.033}                                                   & \multicolumn{1}{c|}{-0.031}                                                    \\ \cline{2-5} 
    \multicolumn{1}{r|}{Noise Visibility$\downarrow$}            & \multicolumn{1}{c|}{30.67}               & \multicolumn{1}{c|}{26.52}                                              & \multicolumn{1}{c|}{\textbf{25.97}}                                           & \multicolumn{1}{c|}{26.49}                                                      \\ \cline{2-5} 
    \multicolumn{1}{r|}{Global Temporal Incoherence$\downarrow$} & \multicolumn{1}{c|}{$2\text{e}^{-5}$}      & \multicolumn{1}{c|}{$2.4\text{e}^{-4}$}                                     & \multicolumn{1}{c|}{$8.6\text{e}^{-4}$}                                       & \multicolumn{1}{c|}{\textbf{$\mathbf{5.7e^{-6}}$}}                                                    \\ \cline{2-5} 
    \multicolumn{1}{r|}{Local Temporal Incoherence$\downarrow$}  & \multicolumn{1}{c|}{0.025}   & \multicolumn{1}{c|}{0.0229}                                      & \multicolumn{1}{c|}{\textbf{0.0207}}                                     & \multicolumn{1}{c|}{0.0258}                                                     \\ \cline{2-5} 
    \end{tabular}
    \label{tab:Denoising_and_Deflickering}
\end{table}

As a summary of this section, we can state that the introduction of the simple temporal regularization loss term $\mathcal{L}_\textit{reg}$ sufficiently handles the flickering issue. The loss term can be easily added to the total loss function and thus becomes part of the joint learning process for denoising and tone mapping. A promising approach for implicitly learning deflickering without the need to use consecutive images during training is proposed by Eilertsen et al.~\cite{Eilertsen2019} and Zheng et al.\cite{Zheng2016} by using data augmentation. Instead of using previous video frames, the training image itself can be augmented by artificially changing its global brightness and thus introducing synthetic flickering. This method is highly promising and should further simplify the training process, but it was not tested in the work presented here.

\section{Comparison with the State-of-the-Art}
\label{sec:sota}

\begin{figure}[htbp]
\centering
\includegraphics[width=1\textwidth]{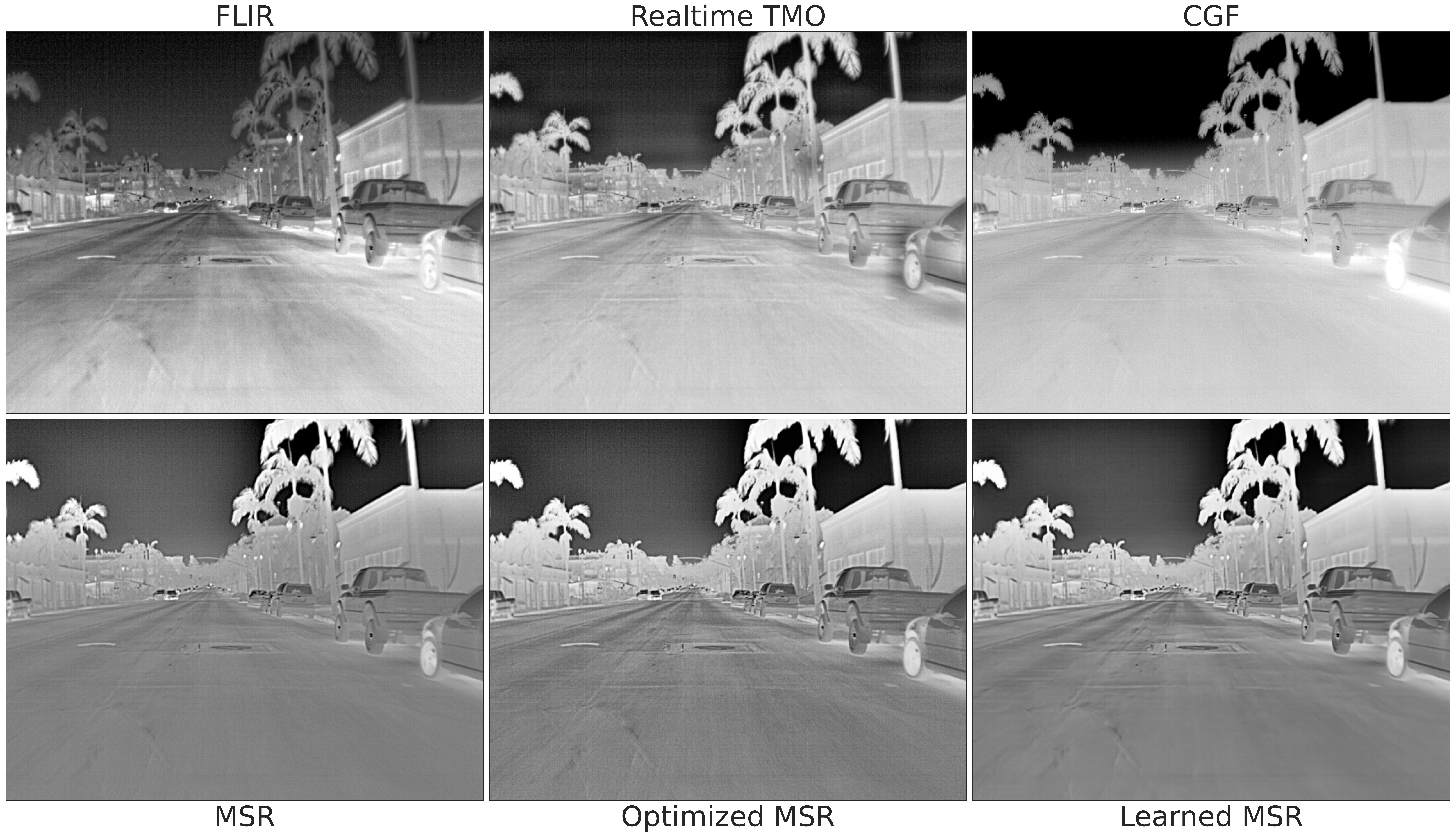}
\caption{Qualitative comparison between the state-of-the-art, our optimized \ac{MSR} and our final deep learning-based approach. The example images confirm the measures in Table~\ref{tab:Final_comparison}: the FLIR algorithm (upper left) has the best TMQI followed by our learned MSR. The noise reduction between the optimized and the learned MSR (bottom center and right) is clearly visible in zoom view.}
\label{fig:SOTA}
\end{figure}

A quantitative evaluation of the main approaches analyzed in this paper is given in Table~\ref{tab:Final_comparison}. The related qualitative comparison is shown in Figure~\ref{fig:SOTA}. Besides the state-of-the-art approaches FLIR, Realtime TMO, \ac{CGF}, and baseline \ac{MSR}, we also include the optimized \ac{MSR} as presented in Section~\ref{sec:retinex} as well as the final deep learning-based approach with joint training of tone mapping, denoising and deflickering as discussed in Section~\ref{sec:TMDCNN}. The optimization of the \ac{MSR} leads to a noticeable improvement in \ac{TMQI}, image contrast, and temporal coherence of the video stream, but also to an increase in noise visibility. Those results can also be confirmed by a qualitative comparison of the images and the video stream. The approximation of the optimized \ac{MSR} through a \ac{DCNN} is called \emph{Learned MSR} in Table~\ref{tab:Final_comparison}. It is able to mimic the image appearance of the optimized \ac{MSR} as we can see in the \ac{TMQI}. After the injection of denoising and deflickering into the training process to jointly learn tone mapping and image enhancement, both the noise visibility and the global temporal incoherence measures clearly indicate an improved image and video quality. We thus can conclude that both optimizations work well together and result in a notable improvement in \ac{TMQI}, noise visibility, and temporal coherence compared to the optimized \ac{MSR}. The contrast measure deteriorated with the joint \ac{DCNN} approach, which, however, could not be confirmed by a qualitative comparison and is perhaps caused by the reduction of noise.

\begin{table}[ht]
    \caption{Comparison between the state-of-the-art, our optimized \ac{MSR} and our deep learning-based approach with joint denoising and deflickering called \emph{learned MSR}. The performance improvement compared to the baseline \ac{MSR}~\cite{Petro2014} is notable in multiple measures with the \ac{TMQI} and the noise visibility being the most important ones. The other Retinex-based approach \ac{CGF} is outperformed, too.}
    \centering
    \begin{tabular}{rcccccc}
    \multicolumn{1}{c}{\textbf{}}   & \textbf{FLIR~\cite{FLIR_dataset}} & \textbf{\begin{tabular}[c]{@{}c@{}}Realtime~\cite{Eilertsen2015}\\TMO\end{tabular}} & \textbf{CGF~\cite{Haibo_conditional_filtering_MSR}} & \textbf{MSR~\cite{Petro2014}} & \textbf{\begin{tabular}[c]{@{}c@{}}Optimized\\MSR\end{tabular}} & \textbf{\begin{tabular}[c]{@{}c@{}}Learned\\MSR\end{tabular}} \\ \cline{2-7} 
    \multicolumn{1}{r|}{TMQI$\uparrow$}  & \multicolumn{1}{c|}{\textbf{0.951}}   & \multicolumn{1}{c|}{0.809}  & \multicolumn{1}{c|}{0.850}  & \multicolumn{1}{c||}{0.866} & \multicolumn{1}{c|}{0.923}  & \multicolumn{1}{c|}{0.934} \\ \cline{2-7} 
    \multicolumn{1}{r|}{Underexposure$\downarrow$} & \multicolumn{1}{c|}{\textbf{0.535}}    & \multicolumn{1}{c|}{0.755}  & \multicolumn{1}{c|}{2.842} & \multicolumn{1}{c||}{1.191}   & \multicolumn{1}{c|}{1.803}  & \multicolumn{1}{c|}{1.330}  \\ \cline{2-7} 
    \multicolumn{1}{r|}{Overexposure$\downarrow$}  & \multicolumn{1}{c|}{0.899}    & \multicolumn{1}{c|}{1.628}  & \multicolumn{1}{c|}{\textbf{0.030}}   & \multicolumn{1}{c||}{1.578} & \multicolumn{1}{c|}{1.176}    & \multicolumn{1}{c|}{1.110} \\ \cline{2-7} 
    \multicolumn{1}{r|}{Loss of Global Contrast$\downarrow$}  & \multicolumn{1}{c|}{-0.162}   & \multicolumn{1}{c|}{-0.123} & \multicolumn{1}{c|}{-0.084}   & \multicolumn{1}{c||}{-0.148}  & \multicolumn{1}{c|}{\textbf{-0.202}}   & \multicolumn{1}{c|}{-0.152} \\ \cline{2-7} 
    \multicolumn{1}{r|}{Loss of Local Contrast$\downarrow$}    & \multicolumn{1}{c|}{-0.041} & \multicolumn{1}{c|}{-0.024} & \multicolumn{1}{c|}{-0.022}   & \multicolumn{1}{c||}{-0.027}   & \multicolumn{1}{c|}{\textbf{-0.044}}  & \multicolumn{1}{c|}{-0.031} \\ \cline{2-7} 
    \multicolumn{1}{r|}{Noise Visibility$\downarrow$}  & \multicolumn{1}{c|}{-}                 & \multicolumn{1}{c|}{30.77}   & \multicolumn{1}{c|}{30.04}  & \multicolumn{1}{c||}{28.79}   & \multicolumn{1}{c|}{30.67}  & \multicolumn{1}{c|}{\textbf{26.49}}  \\ \cline{2-7} 
    \multicolumn{1}{r|}{Global Temporal Incoherence$\downarrow$} & \multicolumn{1}{c|}{$1.7\text{e}^{-4}$}  & \multicolumn{1}{c|}{$3.7\text{e}^{-4}$} & \multicolumn{1}{c|}{$2.8\text{e}^{-5}$} & \multicolumn{1}{c||}{$2.7\text{e}^{-4}$}   & \multicolumn{1}{c|}{$2\text{e}^{-5}$}   & \multicolumn{1}{c|}{\textbf{$\mathbf{5.7e^{-6}}$}} \\ \cline{2-7} 
    \multicolumn{1}{r|}{Local Temporal Incoherence$\downarrow$} & \multicolumn{1}{c|}{\textbf{0.0048}}  & \multicolumn{1}{c|}{0.0157} & \multicolumn{1}{c|}{0.0088} & \multicolumn{1}{c||}{0.0394}  & \multicolumn{1}{c|}{0.025}  & \multicolumn{1}{c|}{0.0258}  \\ \cline{2-7} 
    \end{tabular}
    \label{tab:Final_comparison}
\end{table}

\section{CONCLUSIONS}
\label{sec:conclu}
In this paper, we discussed the topic of tone mapping and image quality enhancement for \ac{TIR} images and videos. Inspired by a promising Retinex-based tone mapping approach for \ac{TIR} images, we set a baseline by the utilization and optimization of an \ac{MSR} algorithm. The optimization mainly focused on contrast enhancement and deflickering. Image noise, however, was amplified. This optimized \ac{MSR} algorithm was then approximated with the popular deep neural network architecture U-Net. Therefore, the images generated by the optimized \ac{MSR} algorithm served as training images in a fully supervised learning scheme. The motivation was to efficiently inject methods into the training approach that allow us to jointly perform denoising and deflickering during training. This was achieved by utilizing a self-supervised image denoising approach based on data augmentation and a simple yet effective deflickering method by introducing a temporal regularization loss term. Extensive experiments were conducted for each step mentioned before. We used the publicly available FLIR ADAS Dataset and the public code of an evaluation framework for tone mapping in the \ac{TIR} spectrum. In a quantitative and qualitative evaluation, we were able to empirically prove the effectiveness of our proposed approaches and demonstrate near state-of-the-art performance in comparison with four tone mapping algorithms taken from the literature. The two Retinex-based tone mapping algorithms were clearly outperformed in this evaluation.

\bibliography{bibliography} 
\bibliographystyle{spiebib} 

\end{document}